# PARAMETRIC CLASSIFICATION OF HANDVEIN PATTERNS BASED ON TEXTURE FEATURES

Harbi AlMahafzah[1,a], Mohammad Imran[2, b] and Supreetha Gowda H D[2, c]

[1]*H.Al-Mahafzah Dept.of Computer Science, College of Information Technology, Al-Hussein Bin Talal University, Ma'an, Jordan)*

[3]*Department of Computer Science, University of Mysore, MYSORE-560 007, India.*

[a]Corresponding author: hmahafzah@hotmail.com
[b]emraangi@gmail.com
[c]supreethad3832@gmail.com

**Abstract:** In this paper, we have developed Biometric recognition system adopting hand based modality Handvein,which has the unique pattern for each individual and it is impossible to counterfeit and fabricate as it is an internal feature. We have opted in choosing feature extraction algorithms such as LBP-visual descriptor,LPQ- blur insensitive texture operator, Log-Gabor-Texture descriptor. We have chosen well known classifiers such as KNN and SVM for classification. We have experimented and tabulated results of single algorithm recognition rate for Handvein under different distance measures and kernel options. The feature level fusion is carried out which increased the performance level.

## INTRODUCTION

With the propagation and expansibility of a network and the anxiety for securing the data from illicit access, demand for personal identification system has increased. The conventional ways (password, id card,..) for authenticating user in accessing to system are no longer safe. With the spring of biometrics technique the conventional authentication systems have been replaced. Biometric automated recognition (1:N) or verification (1:1) system could be developed by physiological (face, fingerprint, iris, retina etc.) and behavioral (signature, gait, lip movement etc.) traits. A simple Biometric module consists of sensor module- acquire the biometric data, Feature Extraction module- in extracting dominant features, matching module to measure the distance between template and the extracted features of the query image, decision making to accept or reject the claimed identity. Biometric characteristics are universality, uniqueness, permanence, collectability, acceptability. Biometrics has the capability to recognize the right identity of different users that entreat services.

In this research we have proposed a biometrics system using hand vein which is inside feature of the body. Finger and palm based system was admit in 2007 by ISO where the vascular biometrics images was standardized.

## RELATED WORK

*Naoto Miura, et al*.[1] proposed a biometric identification system with finger-vein patterns from tenebrousimage using line track. This approach showed that experimental results achieves robust system and lowerEER. Based on the method it can be fused with other hand-biometric based traits.

*V Prasathkumar, and P.Jeevitha*.[2] showed a new way of extracting the features from hand vein modality. The key was used to characterize the vein arrangement i.e., length of the master vein and the position of bifurcation points was measured. They used Hough transform and KNN matching algorithm.

*James ZeWang y et al*.[3] used WBIIS, which characterizes the color divergence to provide semanticallysignificant image comparisons. The Daubechies (DB) wavelet applies the transformation at each component of color image. From the experiments found that, WBIIS performance is higher and more precise than traditional algorithms.

*Sang-KyunImet al*.[4] . extracted the vein DSP processor, which adopts a shift-and-add architecture tofilters. Their experimental result elucidate a good performance of FAR performance without expensive DSP/PC.

*AycanYukselet al*.[5] In their paper, the authors shows a novel hand vein database and a technique based on the statistical handling of the hand vein patterns. They showed a new way of the handvein patterns means of measuring the Line-Segment Hausdorff Distance, using thermal image vein in the back of the hand.

*Lingyu Wang and Graham Leedham*.[6] presented a new approach of person authentication using triangulation of hand vein images and at the same time extraction of knuckle shape. Their method is fully automated and utilize palm dorsal vein shape obtained from the low-cost, near infrared, contactless imaging. According to the authors the experimental results are promising and more user friendly for user identification.

*Lu Yang Gongpinget al*.[7] In their paper, they have adopted support vector machines (SVMs) and they have extracted features such as gradient, contrast, and information capacity from image. A training SVM model on the images is built with quality labels and they applied the method to unseen images to evaluate quality. For the verification of stability on learned model a cross-validation is employed. According to the authors the results show that their method has effectiveness in quality images, by dismissing low-quality images, the aggregate performance recognition is considerably improved.

Kang RyoungPark.[8] In the paper the author proposed a finger vein recognition algorithm, the authors used LBP to gain the local information of fingervein. SVM was adopted to fuse the score values by using LBP and Wavelet transform. The result, shows that the EER was 0.011% low rate.

## PROPOSED METHOD

In this research work, we have analyzed hand based biometric trait such as handvein patterns which refers to image subcutaneous patterns. Since the veins pattern of hand are underneath of the skin will be much difficult to spoof, hence this makes handvein modality as high level of security. In this approach, we have considered some of feature extract methods namely Log gabor, Local Phase Quantization, Local Binary Patterns, Haar and Daubechies wavelets. All these mentioned feature extraction algorithms basically extract the texture of handvein images, which helps in discriminating of one handvein pattern to another. In other hand we have employed well know classification methods such as K-nearest neighbour(k-nn) and support vector machines(SVM). In K-NN we have adopted K=5, with varying different distance measures, however for SVM used polynomial kernel and RBF.

## FEATURE EXTRACTION

In this section, we describe feature extraction algorithms in details which have been used to extract the features prior before training of a classifier.

### Log-Gabor Filters

Field introduced the Log-Gabor[9] Field introduced Log-Gabor [7] filter and showed its ability to represent images by Gaussian transfer functions, which is more efficient compared with the original Gabor filter. Log-Gabor function has a transfer function given by,

$$G(w) = e^{(-log(w/w_0)^2)/(2(log(k/w_0)^2)} \qquad (1)$$

where $w_0$ is the filter's center frequency. To obtain constant shape ratio $k=w_0$ must also be held constant for different values of $w_0$.

## Local Phase Quantization

The LPQ by Ojansivu et al. [10] it is the blur constancy property of the Fourier phase spectrum. It extracted the local phase using the 2-D DFT computed on a rectangular M-by-M neighborhood $N_x$ at each pixel position $x$ of the image $f(x)$ defined by,

$$F(u,x) = \sum_{y \in N_x} f(x-y) e^{-j2\pi u^T y} = w_u^T f_x \qquad (2)$$

where $w_u$ is the basis vector of the 2-D DFT at frequency u, and $f_x$ is another vector containing all $M^2$ image from $N_x$

## Local Binary Patterns

LBP[11] is a visual depictor classifier used in computer vision. It is a robust texture classifier; when LBPis joint with the Histogram of oriented gradients (HOG) depictor, it improves the performance.

## Discrete Wavelet Transform

DWT is any wavelet transform for which the wavelets are discretely sampled[12].

## Daubechies Wavelets

Daubechieswavelets,[13] by Ingrid Daubechies, are a family of perpendicular wavelets define a discrete wavelet transform which can be characterized by a highest number of disappearance instants. There is a calibration function (FatherWavelet) with each wavelet, which can generate an vertical multi-resolution analysis.

## Haar wavelet

Haar wavelet by AlfrdHaar, [14] is a succession of rescaled "square-shaped" functions which all together form a wavelet family or basis Haar which use these functions to give an example of an orthonormal system for the space which is integrable functions on the unit interval [0, 1].

The mother function of Haar wavelet's y(t) can be described as:

$$\psi(t) = \begin{cases} 1 & 0 \leq t < \frac{1}{2} \\ -1 & \frac{1}{2} \leq t < 1 \\ 0 & otherwise \end{cases} \qquad (3)$$

Its scaling function $\phi(t)$ described as

$$\phi(t) = \begin{cases} 1 & 0 \leq t < 1 \\ 0 & otherwise \end{cases} \qquad (4)$$

## FEATURE CLASSIFICATION

### KNN Classifier

The KNN [15] method is based on instance learning which has the ability to stores all data points available and based on similarity measurement the new data could be classified. The idea behind the KNN method is to allocate new examples that to be classify to the class to which the plurality of its K nearest neighbors belongs. KNN make no assumptions on the implied data distribution.

### *Euclidean Distance Metric (EU)*

EU is the ultimate frequent way of computing a distance between two objects. It examines the square root of variation between coordinates of a duo of objects. EU is given below[15]:

$$d_{st} = \sqrt{\sum_{n=1}^{\infty} (x_{sj} - y_{tj})^2} \qquad (5)$$

### *City Block Distance Metric (CB)*

CB first deemed by Hermann Minkowski in the 19th century, is a geometric form of in which the habitual metric of Euclidean geometry is substitute by a new metric in which the distance between two points becomes the sum of the absolute differences of two coordinates illustrate by the following equation[15]:

$$d_{st} = \sum_{n=1}^{\infty} |x_{sj} - y_{tj}| \qquad (6)$$

Here *x* and *y* are Euclidean vectors

### *Correlation Distance Metric (CO)*

CO is one minus the sample correlation between points and is illustrate by the following equation[15]:

$$d_{st} = \frac{(x_s - \bar{x}_s)(y_t - \bar{y}_t)'}{\sqrt{(x_s - \bar{x}_s)(x_s - \bar{x}_s)'}\sqrt{(y_t - \bar{y}_t)'(y_t - \bar{y}_t)}} \qquad (7)$$

where

$$\bar{x}_s = \frac{1}{n}\sum_i x_{sj} \quad and \quad \bar{y}_s = \frac{1}{n}\sum_i y_{sj} \qquad (8)$$

Here *x* and *y* are Euclidean vectors

It is remarkable to point out that the performance of classifiers is generally dependent upon the value of *K*(samples) and distance metric.

## Support Vector Machine

The other classifier used is SVM [16]. for our Experimentation purposes. The concept of SVMs is to make a (nonlinear) mapping function to transform data from input space to data in feature space such that we could achieve linearly separablity. It is a supervised support vector machines used for regression analysis and classification. Given a set of training, each set marked as pertinence to one or other of two categories, an SVM training algorithm construct a sample that assigns to one category or the other, which making a nonprobabilistic binary linear classifier.

Given a set of training, each set marked as pertinence to one or other of two categories, an SVM training algorithm construct a sample that assigns to one category or the other, which making a non-probabilistic binary linear classifier.

### *Gaussian ( Radial Basis Function Kernel) RBF kernel.*

RBF is a prevalent kernel function used in several kernelized learning algorithms. Specifically, it is widely used in support vector machine classification[17].

$$k(x, x') = exp\left(-\frac{||x - x'||^2}{2\sigma^2}\right) \qquad (9)$$

$||x - x'||^2$ known as the squared Euclidean distance between the two feature vectors. s is a free parameter.

*Polynomial kernel*

Polynomial kernel, is a function generally used with support vector machines (SVMs). The following definition is for degree-d polynomials kernel[19]:

$$K(x, y) = (x^T y + c)^d \qquad (10)$$

Where:
$x$ and $y$ are vectors in the input space "a vectors of features of training or test samples".
$c \geq 0$ is a free parameter trading off the leverage of higher-order versus lower-order phrase in the polynomial, homogeneous kernel is establish when $c = 0$.

## RESULTS AND DISCUSSION

This section describes the experiments result evaluate the recognition we obtained from the hand-vein extraction algorithm. First we extracted the features using different algorithm as described above. Next we applied KNN SVM classifiers with different parameters. Finally, we evaluate the fusion of the algorithm using zscore normalizer. Finally, we evaluate the fusion of the algorithm using z-score normalizer.

## One Algorithm With One Classifier

Applying different algorithm with different classifiers along with their parameters.

*KNN Classifier With Different Algorithm*

Table-1 shows the result obtained by different algorithm with K-NN classifier.

**TABLE 1.** Performance of K-NN Classifier for Handvein

| Distance Measure | Recognition rate (%) | | | | |
| --- | --- | --- | --- | --- | --- |
| | LBP | LPQ | Gabor | Db8 | Haar |
| Euclidean | 64.50 | 70.20 | 66.50 | 60.50 | 71.50 |
| City block | 65.00 | 72.00 | 69.50 | 64.50 | 70.50 |
| Cosine | 62.00 | 71.00 | 67.00 | 59.00 | 69.00 |
| Correlation | 62.50 | 69.50 | 67.50 | 57.50 | 67.50 |

Table-1 shows the recognition rate for Handvein modality obtained from well-known feature extraction algorithms such as Local Binary Patterns, Local Phase Quantization, Daubechies 8 wavelet, Haar wavelets. The features are well trained with K-NN classifier and the nearest neighbors can be calculated using Euclidean, City block, Cosine, Correlation distances , as it gives good productivity and the results are evaluated. One can observe that, LPQ and HAAR feature extraction algorithms is yielding good recognition rate for the KNN classifier implemented on all the considered distance measures. Db8 is underperforming when compared with other feature extractors and we can't rely on this technique for good recognition rate.

*SVM classifier With different Algorithm*

Table-2 shows the result obtained by different algorithm with SVM classifier.

**TABLE 2.** Performance of SVM Classifier for Handvein

| Distance Measure | Recognition rate (%) | | | | |
|---|---|---|---|---|---|
| | LBP | LPQ | Gabor | Db8 | Haar |
| Polynomial | 66.50 | 76.50 | 72.50 | 62.00 | 73.50 |
| Gaussian | 73.50 | 79.00 | 75.00 | 64.50 | 76.50 |

The features are extracted from LBP, LPQ, Gabor, Haar, Db8 feature extraction techniques, the features are trained with Support Vector Machine Classifier- A supervised learning and a statistical approach which handles curse of dimensionality. Choosing a suitable kernel for SVM classifier depends on application. The most common kernel options used are Polynomial and Gaussian kernel, which we have employed in our experimentation. Separation of non-linear data of high dimension to linearly separable data truly depends on data set.

In our case, from the obtained results we can say that Gaussian kernel has very good impact than Polynomial kernel with the classification rate obtained from the Performance of SVM Classifier. Here also LPQ is providing outstanding classification rate of 76:5%, 79% under the polynomial and Gaussian kernels respectively, HAAR is also providing considerable recognition rate when compared with other techniques

## Two Algorithms With One Classifier

Z-score normalization rule used to fuse the two algorithm used with one classifiers. The z-score normalized score is given by: $ns_j^t = \frac{s_j^t - \mu_j}{\sigma_j}$

*LPQ and Haar wavelet with K-NN Classifier*

Table-3 shows the performance of K-NN Classifier for feature level fusion of LPQ and Haar using Z-score normalization rule.

**TABLE 3.** Performance of K-NN Classifier for feature level fusion of LPQ and Haar using Z-score normalization rule.

| Distance Measure | LPQ + Haar | | | |
|---|---|---|---|---|
| | Fusion Recognition rate (%) | Recall | Precision | F-measure |
| Euclidean | 84.5 | 0.845 | 0.848 | 0.846 |
| City Block | 81.0 | 0.810 | 0.832 | 0.821 |
| Cosine | 85.0 | 0.850 | 0.847 | 0.848 |
| Correlation | 80.0 | 0.800 | 0.779 | 0.789 |

From Table-1 and Table-2 we can see that, the best recognition is achieved from LPQ and Haar techniques, so we have performed pre classification-feature level fusion of these two methods. The feature vectors obtained from LPQ and Haar are concatenated in getting a single discriminating feature vector. The feature vectors obtained from HAAR wavelets are normalized using Z-score measure for attaining homogeneity in fusing feature vectors. The performance is measured by computing Recall, Precision, F-Measure and the recognition rate is computed and the K-NN classifier which is implemented using the distance measures such as Euclidean, City Block, Cosine and Correlation. From the results obtained, we can say that Cosine and Euclidean based distance measures is giving good classification rate of 85%;84:5% respectively. We can also infer that feature level fusion has better performance when compared with Table-1 results.

*LPQ and Haar wavelet with SVM Classifier*

Table-4 shows the Performance of SVM Classifier for feature level fusion of LPQ and Haar using Z-score normalization rule.

TABLE 3. Performance of SVM Classifier for feature level fusion of LPQ and Haar

| Kernel option | LPQ + Haar | | | |
|---|---|---|---|---|
| | Fusion Recognition rate (%) | Recall | Precision | F-measure |
| Polynomial | 90.5 | 0.905 | 0.898 | 0.904 |
| Gaussian | 89.0 | 0.890 | 0.891 | 0.890 |

Table-4 shows the classification rate obtained from SVM classifier under feature level fusion of LPQ and Haar feature extractors. The performance measures like Recall, precision, F-Measure has gained good results and the recognition rate has increased in adopting SVM classifier when compared to K-NN classifier. Polynomial kernel is the good kernel option than the Gaussian kernel, as it has achieved 90:5% of fusion recognition rate.

## CONCLUSIONS

Analyzing the recognition rate using hand vein images as a biometric has been done. The results of evaluation represented by means of system recognition rate (expressed %) Experimental results tabulatedinfers the comparison between one algorithm and feature level fusion under multi-algorithms biometric identification system for the Handvein modality, SVM classifier is performing better than the KNN classifier in both one algorithm and multi-algorithms biometric recognition system.

Multi- Biometric system has higher recognition rate when compared with single algorithm system. The various limitations of the uni-biometrics system is usually addressed by multi-biometrics system by means of combining various features. However, from the experimental results and observations the degree of improvement in accuracy by fusing multiple instances is notable. Our future work would be intended in carrying out research with behavioral traits and concentrating more on feature extraction algorithms for the effective classification.